\documentclass[conference]{IEEEtran}
\IEEEoverridecommandlockouts

\usepackage{algorithm}
\usepackage{algpseudocode}
\usepackage{bbold}
\usepackage{booktabs}
\usepackage{caption}
\usepackage{amsmath,amssymb,amsfonts}
\usepackage{enumitem}
\usepackage{float}
\usepackage{graphicx}
\usepackage{makecell}
\usepackage{multirow}  
\usepackage{multicol}
\usepackage{pgfplots}
\usepackage{lipsum}
\usepackage{subcaption}
\usepackage{textcomp}
\usepackage{tikz,tikzscale}
\usepackage{xcolor}
\usepackage{adjustbox}

\def\BibTeX{{\rm B\kern-.05em{\sc i\kern-.025em b}\kern-.08em
    T\kern-.1667em\lower.7ex\hbox{E}\kern-.125emX}}

\algnewcommand\algorithmicinput{\textbf{Input:}}
\algnewcommand\Input{\item[\algorithmicinput]}
\algnewcommand\algorithmicoutput{\textbf{Output:}}
\algnewcommand\Output{\item[\algorithmicoutput]}
\begin{document}

\title{Towards a Larger Model via One-Shot Federated Learning on Heterogeneous Client Models 
}

\author{\IEEEauthorblockN{Wenxuan Ye\IEEEauthorrefmark{1}\IEEEauthorrefmark{2},
Xueli An\IEEEauthorrefmark{1}, 
Onur Ayan\IEEEauthorrefmark{1}, 
Junfan Wang\IEEEauthorrefmark{3},
Xueqiang Yan\IEEEauthorrefmark{3},
Georg Carle\IEEEauthorrefmark{2}}
\IEEEauthorblockA{
\IEEEauthorrefmark{1} Advanced Wireless Technology Laboratory, Munich Research Center, Huawei Technologies Duesseldorf GmbH\\
\IEEEauthorrefmark{2} TUM School of Computation, Information and Technology, Technical University of Munich\\
\IEEEauthorrefmark{3} Wireless Technology Lab, 2012 Laboratories, Huawei Technologies Co., Ltd\\
wenxuan.ye@tum.de, \{xueli.an, onur.ayan, wangjunfan3, yanxueqiang1\}@huawei.com, carle@net.in.tum.de
}}

\maketitle

\begin{abstract}
Large models, renowned for superior performance, outperform smaller ones even without billion-parameter scales.
While mobile network servers have ample computational resources to support larger models than client devices, privacy constraints prevent clients from directly sharing their raw data.
Federated Learning (FL) enables decentralized clients to collaboratively train a shared model by exchanging model parameters instead of transmitting raw data.
Yet, it requires a uniform model architecture and multiple communication rounds, which neglect resource heterogeneity, impose heavy computational demands on clients, and increase communication overhead.
To address these challenges, we propose \textit{FedOL}, to construct a larger and more comprehensive server model in one-shot settings (i.e., in a single communication round). 
Instead of model parameter sharing, FedOL employs knowledge distillation, where clients only exchange model prediction outputs on an unlabeled public dataset.
This reduces communication overhead by transmitting compact predictions instead of full model weights and enables model customization by allowing heterogeneous model architectures.
A key challenge in this setting is that client predictions may be biased due to skewed local data distributions, and the lack of ground-truth labels in the public dataset further complicates reliable learning. 
To mitigate these issues, FedOL introduces a specialized objective function that iteratively refines pseudo-labels and the server model, improving learning reliability.
To complement this, FedOL incorporates a tailored pseudo-label generation and knowledge distillation strategy that effectively integrates diverse knowledge.
Simulation results show that FedOL significantly outperforms existing baselines, offering a cost-effective solution for mobile networks where clients possess valuable private data but limited computational resources.
\end{abstract}


\section{INTRODUCTION}
The introduction of ChatGPT and Deepseek \cite{Brown2020, Guo2025} has significantly raised awareness of the potential in large-scale models, which offer more accurate predictions, better decision-making, and advanced automation.
Research indicates that larger, more complex models generally outperform smaller ones, even without reaching the billion-parameter scale \cite{Brutzkus2019, Qian2022}.
While mobile network servers have ample computational resources to support such complex models, privacy concerns prevent clients from sharing raw data directly \cite{Kairouz2021}. 
Federated Learning (FL) \cite{McMahan2017} addresses this challenge by enabling decentralized clients to collaboratively train a shared model while preserving data privacy. 
In classical FL, clients train local models on their private data and transmit model parameters to the server.
The server then updates its model by averaging the received parameter \cite{McMahan2017} or related methods \cite{Li2020}, and broadcasts the updated model to the clients.

Nonetheless, several challenges hinder the application of FL in developing larger models.
First, such \textit{parameter-based sharing paradigms} assume \textit{uniform model architectures} across participants, overlooking resource heterogeneity and leading to significant performance disparities among client models. 
Second, FL typically relies on \textit{multiple communication rounds} with clients, necessitating tightly synchronized information transmission, extensive client-side computation and communication overhead and risking privacy leakage \cite{Kairouz2021, Ye2022}.
Third, \textit{diverse client data distribution} introduces further complexity, increasing the risk of model overfitting in localized data domains and failing to achieve broad generalizability \cite{Ye2023}.

Instead of sharing model parameters, Knowledge Distillation (KD) offers an alternative by exchanging model predictions over a public dataset.
In KD, a student model learns to mimic the prediction probabilities of a pre-trained teacher model on the shared dataset \cite{Hinton2015}.
Applied to FL scenarios, client models act as teachers, providing prediction outputs that guide the server in constructing a larger, more generalized model \cite{Cho2022}. 
This method enables model customization and reduces communication overhead compared to traditional parameter-based approaches.
While some methods rely on labeled datasets \cite{Cheng2021}, others (e.g., FedDF \cite{Lin2020}) employ unlabeled data to bypass the costly and time-consuming labeling.

To address the limitation of uniform model architecture and multiple communication rounds in traditional FL, we propose \textit{FedOL}—\textit{a one-shot FL approach} where clients share their knowledge \textit{only once} during training. 
Rather than exchanging model parameters, FedOL utilizes KD and collects \textit{client prediction outputs on an unlabeled public dataset}, enabling the server to build a larger global model.
By restricting client participation to a single round of local training and prediction sharing, FedOL reduces computation and communication overhead, and aligns with modern trends where clients contribute pre-trained models instead of engaging in iterative updates \cite{Kairouz2021}.

One key challenge in this setting is the prevalence of data heterogeneity in real-world scenarios, which leads to biased client predictions, an issue further complicated by the lack of ground truth in public datasets for calibration.
Although client heterogeneity introduces challenges, it also presents an opportunity to leverage the specialized strengths of individual models.
Building on this insight, FedOL introduces \textit{a specialized objective function} that iteratively refines pseudo-labels and updates the server model, improving learning reliability.
To complement this, FedOL incorporates \textit{a tailored pseudo-label generation and KD strategy} that effectively integrates diverse client knowledge.

We conduct extensive experiments on CIFAR-100, comparing FedOL with baselines across diverse data distributions. 
Our evaluation includes a fair comparison where all the methods are tested under one communication round, as well as baseline simulations across multiple rounds, and detailed cost analysis. 
Results highlight that FedOL delivers high performance while minimizing client communication and computation, offering a significant advantage in cost-sensitive scenarios.

We highlight the contributions of this paper as follows:
\begin{itemize}[left=0em]
    \item {\textbf{System design:} 
    We propose FedOL, a one-shot FL framework that constructs a large and comprehensive server model by collecting a single round of client model predictions on an unlabeled public dataset. 
    It minimizes client-side computation and communication by simplifying client involvement to one-round local training and then prediction sharing.
    }
    \item {\textbf{Algorithm optimization:} 
    We introduce a specialized objective function that jointly refines pseudo-labels and improves the server model through iterative optimization.
    To address client data heterogeneity, we further incorporate a tailored pseudo-label generation and KD strategy.
    }
    \item {\textbf{Simulation:} 
    Extensive experiments show that FedOL outperforms baselines in both cost-efficiency and accuracy, particularly under heterogeneous data distributions.
    }
\end{itemize}

\section{BACKGROUND AND RELATED WORK}

\noindent \textbf{Knowledge Distillation in Federated Learning:}
In addition to the \textit{label loss} that aligns the server model’s predictions with (pseudo) labels, KD introduces a \textit{distillation loss}, encouraging the server model $w_s$ to mimic the prediction outputs of client models ${w_c^k}$ on an unlabelled shared dataset $\mathbb{D}_u$ \cite{Hinton2015}.
The server aggregates the output logits from client models, either uniformly as in \cite{Lin2020} or with varying weights \cite{Cho2022}.
The aggregated knowledge guides the training of the server model by the following optimization problem:
\begin{equation}
\min_{w_s} \mathbb{E}_{x \in \mathbb{D}_u} \left[\text{D}_{\text{KL}}\left(\sigma(\sum\nolimits_{k}{\mu_{k} w_c^k(x)}) \: ||\: \sigma(w_s(x))\right)\right] 
\end{equation}
where $\text{D}_{\text{KL}}(\cdot)$ represents the Kullback-Leibler Divergence, $\sigma(w(x))$ for the softmax output of model $w$ evaluated at input $x$, $\mu_{k}$ for the weight assigned to client $k$, with $\sum_{k}{\mu_k} = 1$.

In KD, many approaches improve performance by only sharing client predictions that meet certain confidence criteria, including confidence-based thresholds \cite{Li2022} or additional regularization techniques \cite{Cho2021}.
To better address data heterogeneity, recent work has adopted adaptive aggregation techniques, such as clustering similar clients \cite{Deng2023}, or analyzing variance in model predictions \cite{Cho2022}.
However, these methods typically rely on multiple communication rounds, increasing the burden on client resources, exposing potent security risks, and ignoring each client model's unique contributions.

\noindent \textbf{One-shot Federated Learning:}
One-shot FL, first introduced in \cite{Guha2019}, restricts communication between clients and the server to a single round. 
Methods like DENSE \cite{Zhang2022} achieve this by generating synthetic data from client models, but raise concerns around privacy and client-side computation.
Alternatively, FedOV \cite{Diao2023} improves performance through algorithmic enhancements, generating diverse outliers and introducing an unknown class during local training. 
However, its reliance on data generation still imposes significant computational overhead on clients.
In contrast, our method, FedOL, leverages client-shared prediction outputs to optimize the server-side aggregation algorithm, effectively reducing client computation and preserving privacy without synthetic data generation.

\section{SYSTEM AND PROBLEM DESCRIPTION}

\subsection{System Overview}

\begin{figure}[t!]
    \begin{center}
    \includegraphics[width=\columnwidth]{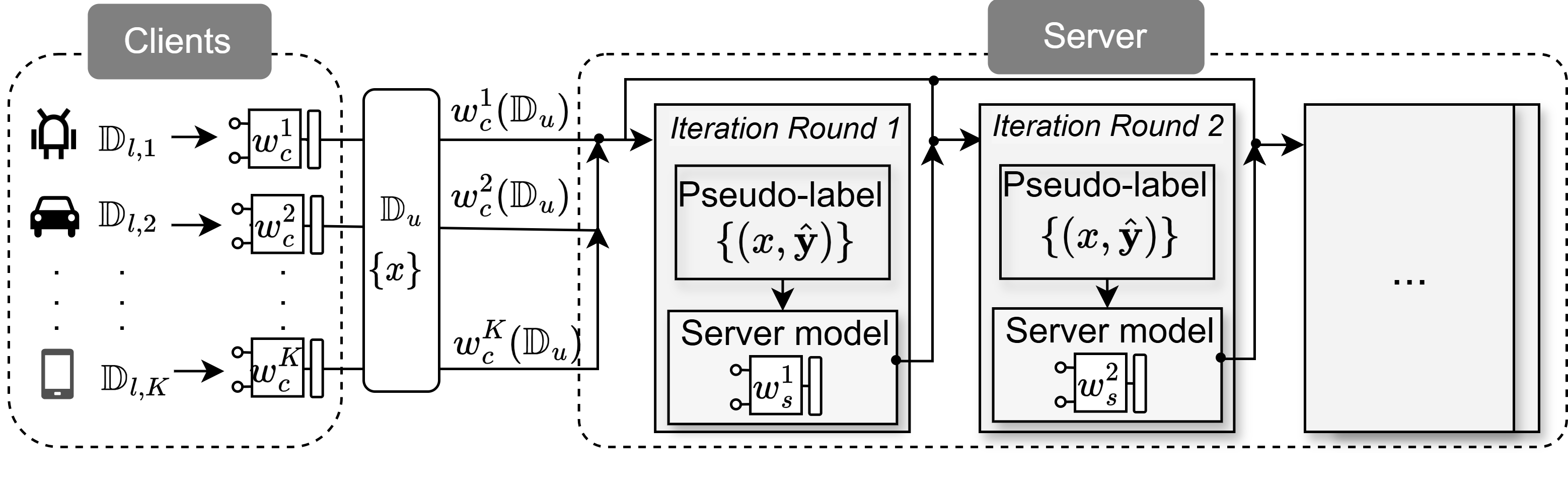}
    \captionsetup{font=small}
    \caption{FedOL System. 
    All participants share a public unlabeled dataset $\mathbb{D}_{u}$. 
    Each client $k$ keeps their private labeled data $\mathbb{D}_{l,k}$ and model $w_c^{k}$ confidential, sharing only the predictions $w_c^k(\mathbb{D}_{u})$ with the server. 
    The server then employs an iterative optimization process on pseudo-label generation and model updates, to train $w_s^{t}$, where $w_s^{t}$ denotes the server model at the $t$-th iteration.
    }
    \label{fig:arch}
    \end{center}
\end{figure}

Fig.~\ref{fig:arch} illustrates our system, which comprises $K$ clients.
Each client $k$ maintains a private labeled dataset denoted by $\mathbb{D}_{l,k} = \{(x, \mathbf{y})\}$, where $x$ represents the data sample, and $\mathbf{y}$ its one-hot truth label.
Although each client’s dataset may have a different distribution, all clients share the same classification task with $C$ classes.
We denote the complete labeled dataset as $\mathbb{D}_{l} = \cup_{k = 1}^{K} \mathbb{D}_{l,k}$.
Additionally, all participants share a public unlabeled dataset $\mathbb{D}_{u} = \{x\}$, which serves as the medium for knowledge transfer.
We adopt an unlabeled dataset for its wide availability and to avoid the significant cost and effort of labeling.
The system's objective is to develop a larger and more comprehensive server-side model $w_s$, without exposing clients' raw data or model parameters.

Since clients only share model predictions, they are free to customize their model architecture, such as network type and size, according to their computational capacity.
Each client $k$ independently trains its local model $w_c^{k}$ on its private dataset $\mathbb{D}_{l,k}$, by optimizing the loss:
\begin{equation}
\label{eq:local}
w_c^k = \arg \min_{w} \mathbb{E}_{(x, \mathbf{y}) \in \mathbb{D}_{l,k}} [\text{CE}\left(\mathbf{y}, \sigma(w(x))\right)]
\end{equation}
where $\text{CE}(\cdot)$ is the cross-entropy function, defined as $\text{CE}(\mathbf{y}, \textbf{p}) = -\mathbf{y}^T \log \textbf{p}$ for labels $\mathbf{y}$ and model predictions $\textbf{p}$.
After training, each client $k$ shares its predictions on the public dataset, denoted as $w_c^{k}(\mathbb{D}_{u})$.
No further communication or participation is required from the client.

To learn from client predictions, the server performs an iterative process. 
In each iteration round $t$, FedOL generates pseudo-labels from model predictions and updates the server model $w_s^t$, progressively improving performance.

\subsection{Problem Formulation}
\noindent{\textbf{Semi-supervised learning:}}
We begin by reviewing semi-supervised learning, which leverages both labeled and unlabeled data to train models, unlike supervised learning which relies solely on labeled data \cite{Grandvalet2004}.
A common approach first trains a model $w$ on a labeled dataset $\mathbb{D}_l = \{(x, \mathbf{y})\}$, then uses it to generate pseudo-labels for an unlabeled dataset $\mathbb{D}_u = \{x\}$. 
The model is subsequently retrained on the combined set of labeled and pseudo-labeled samples.
The overall loss function is defined as: 
\begin{equation}
\label{eq:ssl}
\begin{aligned}
\mathcal{L}(w, \hat{\mathbf{y}}) = & \underbrace{\mathbb{E}_{(x, \mathbf{y}) \in \mathbb{D}_{l}} \left[\text{CE}(\mathbf{y}, \sigma(w(x)))\right]}_{: \mathcal{L}_{l}(w)}  \\
& + \tau \underbrace{\mathbb{E}_{x \in \mathbb{D}_{u}} \left[\text{CE}(\hat{\mathbf{y}}, \sigma(w(x)))\right]}_{: \mathcal{L}_{u}(w, \hat{\mathbf{y}})}
\end{aligned}
\end{equation}
where $\hat{\mathbf{y}}$ denotes the pseudo-label for the unlabeled sample $x$; $\tau \in \mathbb{R}^{+}$ controls the trade-off between the supervised loss $\mathcal{L}_{l}(w)$ and the pseudo-label loss $\mathcal{L}_{u}(w, \hat{\mathbf{y}})$. 
Typically, $\hat{\mathbf{y}}$ is generated by assigning 1 to the class with the highest predicted probability and 0 to all others, forming a one-hot vector \cite{Grandvalet2004}.
By jointly optimizing $w$ and $\hat{\mathbf{y}}$ to minimize $\mathcal{L}(w, \hat{\mathbf{y}})$, the refined pseudo-labels increasingly approximate the true underlying target labels, improving the model’s efficacy in leveraging unlabeled data.
However, in heterogeneous settings, client data may have imbalanced label distributions, causing models to overfit to frequent classes and overlook rare ones.

\noindent{\textbf{Our design:}}
The entire dataset of our system consists of labeled data $\mathbb{D}_{l}$ and unlabeled data $\mathbb{D}_{u}$, mirroring the framework of semi-supervised learning.
Accordingly, we design our loss function based on (\ref{eq:ssl}).
However, since the labeled data is inaccessible for the server, an approximation is necessary.

Through local training as defined in (\ref{eq:local}), clients are likely to accurately learn the true labels of their own samples.
Using the softmax function $\sigma(\cdot)$ as a surrogate for $\arg \max (\cdot)$, we approximate $\mathcal{L}_{l}(w_s)$ as:
\begin{equation}
\begin{aligned}
\mathcal{L}_l(w_s) & = \mathbb{E}_{k \in [1, K], (x,\mathbf{y}) \in \mathbb{D}_{l,k}} \left[\text{CE}(\mathbf{y}, \sigma(w_s(x)) \right]\\
&\approx  \mathbb{E}_{k \in [1, K], (x,\mathbf{y}) \in \mathbb{D}_{l,k}} \left[\text{CE}(\sigma(w_c^k(x)), \sigma(w_s(x)))\right]
\end{aligned}
\end{equation}
where each truth label $\mathbf{y} \in \mathbb{D}_{l,k}$ is approximated by the local model's predictions $\sigma(w_c^k(x))$.

We recognize that $\text{CE}(\mathbf{u},\mathbf{v}) = \text{D}_{\text{KL}}(\mathbf{u} || \mathbf{v}) + \mathcal{H}(\mathbf{u})$, where $\mathbf{u}$ and $\mathbf{v}$ represent two probability distributions, and $\mathcal{H}(\cdot)$ represents the entropy calculation function.
Since the entropy term $\mathcal{H}(\sigma(w_c^k(x))$ is independent of $w_s$, we can exclude that from the loss function, and define a new function $\mathcal{L}_d(w_s)$:
\begin{equation}
\begin{aligned}
\mathcal{L}_d(w_s) &= \mathcal{L}_l(w_s) - \mathbb{E}_{k \in [1, K], (x,\mathbf{y}) \in \mathbb{D}_{l,k}}[\mathcal{H}(\sigma(w_c^k(x))] \\
                 &= \mathbb{E}_{k \in [1, K], (x,\mathbf{y}) \in \mathbb{D}_{l,k}} [\text{D}_{\text{KL}}\left(\sigma(w_c^k(x)) \: ||\: \sigma(w_s(x))\right)] \\
                 &=  \mathbb{E}_{x \in \mathbb{D}_{l}} \left[ \sum\nolimits_k \mathbb{1}[x \in {\mathbb{D}_{l, k}}] \: \text{D}_{\text{KL}}(\sigma(w_c^k(x)) || \sigma(w_s(x)))\right]\\ 
\end{aligned}
\end{equation}
where $\mathbb{1}[\cdot]$ is used as the binary indicator of the event.

Since direct access to private labeled data ${\mathbb{D}_{l}}$ is infeasible, we approximate this loss over public datasets ${\mathbb{D}_{u}}$.
In this context, in order to approximate $\mathbb{1}[x \in {\mathbb{D}_{l, k}}]$, we introduce a weight $\lambda_{k}(x)$ that measures the similarity between a public sample $x \in \mathbb{D}_{u}$ and the training data $\mathbb{D}_{l, k}$.
Thus, $\mathcal{L}_d(w_s)$ is approximated as:
\begin{equation}
\mathcal{L}_d(w_s) \approx \mathbb{E}_{x \in {\mathbb{D}_u}} \left[\sum\nolimits_{k} \lambda_{k}(x) \text{D}_{\text{KL}}\left(\sigma(w_c^k(x)) \: || \: \sigma(w_s(x))\right)\right]
\end{equation}
The design of $\lambda_{k}(x)$ requires careful consideration, and will be discussed in detail in the subsequent section.

In summary, the overall objective function for the server model $w_s$ is defined:
\begin{equation}
\label{eq:overall}
\begin{aligned}
\mathcal{L}(w_s, \hat{\mathbf{y}})= & \underbrace{\mathbb{E}_{x \in {\mathbb{D}_u}} \left[ \sum\nolimits_{k} \lambda_{k}(x) \: \text{D}_{\text{KL}}(\sigma(w_c^k(x)) || \sigma(w_s(x)))\right]}_{: \mathcal{L}_{d}(w_s)} \\
& + \tau \underbrace{\mathbb{E}_{x \in \mathbb{D}_{u}} \left[\text{CE}(\hat{\mathbf{y}}, \sigma(w_s(x)))\right]}_{: \mathcal{L}_{u}(w_s, \hat{\mathbf{y}})} \\
\text{s.t. } \hat{\mathbf{y}} \in  \{\mathbf{y} | \mathbf{y} \in &\{0, 1\}^C,  \sum\nolimits_{c} y_{c} =1 \}
\end{aligned}
\end{equation}
with $y_{c}$ being the c-th component of $\mathbf{y}$.

We denote $\mathcal{L}_{d}(w_s)$ as the \textit{distillation loss}, and $\mathcal{L}_{u}(w_s, \hat{\mathbf{y}})$ as the \textit{pseudo-label loss}.
It is noteworthy that this loss function bears a resemblance to the one used in KD, which also consists of label loss and distillation loss.

\section{DESIGN OF FEDOL}
\label{sec:algorithm}

\subsection{Design Rationale}
The inherent challenge in optimizing (\ref{eq:overall}) stems from the uncertainty of pseudo-label accuracy, particularly in the early stages of model training where the server model $w_s$ has limited accuracy.
While incorporating client models can be beneficial, the diversity in their data requires a carefully designed pseudo-label generation function to ensure reliable integration of their predictions.
This function is designed to capitalize on the strengths of both the client models $\{w_c^k\}_{k=1}^{K}$ and the server model $w_s$, thereby facilitating a more robust learning process from heterogeneous sources of information. 
We define the pseudo-label generation process for input $x$ as:
\begin{equation}
 \hat{\mathbf{y}} = \phi\left(\mathcal{G}(\{w_c^k(x)\}_{k=1}^{K}, w_s(x))\right)
\end{equation}
where $\phi(\cdot)$ denotes a one-hot encoding function, and we simplify $\mathcal{G}(\{w_c^k(x)\}_{k=1}^{K}, w_s(x))$ to $\mathcal{G}(w_s, x)$ for streaming notation.
Details on $\mathcal{G}$ will be elaborated later.

Drawing inspiration from the ``easy-to-hard'' strategy in self-paced curriculum learning \cite{Jiang2015}, we generate pseudo-labels initially from the most reliable model predictions.
As the server model $w_s$ adapts and its performance improves, the strategy gradually incorporates less confident pseudo-labels.
To enable this adaptive process, we introduce a hyperparameter $\rho \in (0, 1]$ that serves as a confidence threshold.
Pseudo-labels $\hat{\mathbf{y}}$ falling below this threshold are excluded from training by assigning them an all-zero vector $\mathbf{0}$.
The updated labeling function is defined as $\mathcal{G}(w_s, x, \rho)$.
Consequently, the generated pseudo-label is described by the condition:
\begin{equation}
\hat{\mathbf{y}} \in  \{\mathbf{y} | \mathbf{y} \in \{0, 1\}^C,  \sum\nolimits_{c} y_{c} \leq 1 \}
\end{equation}

During the algorithm's weight annealing, we increase $\rho$ to progressively lower this threshold, thereby admitting a broader set of pseudo-labels into training.

\begin{table*}[t!]
\setlength{\abovecaptionskip}{2pt} 
\setlength{\belowcaptionskip}{2pt} 
\centering
\caption{Performance comparison under one communication round}
\label{tab:Diri}
\begin{tabular}{c|c | c |c c |c c c c |c}
\toprule
\textbf{Partition} & \textbf{FedOL} & \textbf{Local} & \textbf{FedAvg} & \textbf{FedProx} 
& \textbf{FedDF} & \textbf{FedHKT} & \textbf{FedET}  & \textbf{MinE} 
 & \textbf{FedKT} \\
\midrule
$p \sim \text{Dir}(1)$ & $\mathbf{37.2_{\pm 1.7}}$ & $20.0_{\pm 2.0}$ & $1.1_{\pm 0.2}$ & $1.7_{\pm 0.6}$ 
& \textit{\underline{$29.8_{\pm 1.4}$}} & $29.4_{\pm 1.6}$ & $25.8_{\pm 0.8}$ & $26.2_{\pm 4.6}$
& $20.5_{\pm 2.7}$ \\
$p \sim \text{Dir}(0.05)$ & $\mathbf{29.8_{\pm 1.3}}$ & $10.3_{\pm 0.3}$ & $1.0_{\pm 0.1}$ & $1.0_{\pm 0.0}$
& $14.8_{\pm 1.4}$ & \textit{\underline{$18.8_{\pm 2.3}$}} & $18.0_{\pm 1.1}$ & $15.5_{\pm 1.0}$
& $13.2_{\pm 1.1}$ \\
\midrule
$\# I_{c} = 30$ & $\mathbf{35.8_{\pm 1.1}}$ & $14.8_{\pm 1.3}$ & $1.1_{\pm 0.1}$ & $1.5_{\pm 0.2}$
& $26.1_{\pm 1.1}$ & $26.2_{\pm 1.3}$ & $25.0_{\pm 1.5}$ & \textit{\underline{$30.0_{\pm 1.0}$}}
& $18.0_{\pm 0.6}$ \\
$\# I_c = 20$ & $\mathbf{33.5_{\pm 2.4}}$ & $11.5_{\pm 1.2}$ & $1.0_{\pm 0.0}$ & $1.1_{\pm 0.2}$
& $21.3_{\pm 2.1}$ & \textit{\underline{$25.2_{\pm 0.5}$}} & $22.2_{\pm 0.7}$ & $17.5_{\pm 1.0}$
& $12.6_{\pm 0.6}$ \\
\bottomrule
\end{tabular}
\end{table*}

\subsection{Algorithm Design}
To minimize (\ref{eq:overall}), we employ an alternating block coordinate algorithm that iterates over two steps:
\begin{enumerate}[label=\alph*)]
    \item \textit{Pseudo-label generation}: Fix $w_s$ and minimize the loss with respect to $\hat{\mathbf{y}}$ using function $\phi(\mathcal{G}(w_s, x, \rho))$.
    \item \textit{Server model training}: Fix $\hat{\mathbf{y}}$ and optimize the objective with respect to $w_s$.
\end{enumerate}

\noindent Each iteration round consists of one execution of step a) followed by step b). 
Next, we detail the implementation aspects of the algorithm.

\noindent\textbf{Class-wise confidence:}
Since each participating model tends to excel at classifying certain categories, we utilize class-wise confidence scores to capture these strengths. 
Inspired by the insight from \cite{Diao2023} that a model's predictions mirror the underlying data distribution, we use these class-wise scores as voting weights when aggregating predictions.
Concretely, for client $k$, we collect the predicted probability $\sigma(w_c^k(x))$ for each unlabeled sample $x \in \mathbb{D}_u$, and then compute its expectation over these probabilities across the entire dataset to generate a \textbf{C}lass-wise \textbf{C}onfidence score $\mathbf{C^2}$:
\begin{equation}
\label{eq:C^2}
\mathbf{C^2}_k = \mathbb{E}_{x \in \mathbb{D}_u}[\sigma(w_c^k(x))] 
\end{equation}
where $\mathbf{C^2}$ is C-dimensional vector.

\noindent\textbf{Pseudo-label generation:}
We propose a voting-based pseudo-label generation scheme that robustly integrates heterogeneous model predictions and alleviates class bias under skewed data distributions. 
At the $t$-th iteration round, we define the set of source models as $\mathbb{M} = \{w_c^k\}_{k=1}^{K} \cup \{w_s^{t-1}\}$, comprising both the client models and the server model. 
The weighted vote from each model is integrated into a collective decision-making framework, with the algorithm detailed in Algorithm~\ref{alg:initial}.
Note that in the first iteration, only client models serve as source models. 
Throughout the process, the server never accesses client models or raw data directly, as all learning is based solely on the predictions shared by clients.

\textit{Step 1}: each model $w_m \in \mathbb{M}$ establishes an entropy baseline $\tilde{H}_m$.
The baseline is set to the $\lceil \rho \: N_u \rceil$-th lowest prediction entropy, where $N_u = |\mathbb{D}_{u}|$. 
This threshold identifies the most confident predictions by including only samples whose entropy is below $\tilde{H}_m$, thus balancing the trade-off between capturing a sufficient number of confident predictions and maintaining robust performance. 
As iteration rounds proceed, the parameter $\rho$ is gradually increased according to the ``easy-to-hard" principle, thereby admitting a larger set of pseudo-labels as the server model improves.

\textit{Step 2}: for each sample $x$, the pseudo-label $\hat{\mathbf{y}}$ is initialized to $\mathbf{0}$. 
We define the set of reliable source models $\mathbb{M}(x)$ as:
\begin{equation}
\mathbb{M}(x) = \{w_m \: |\:  \mathcal{H}(\sigma(w_{m}(x))) \leq \tilde{H}_m\}
\end{equation}
Only models whose prediction entropy on $x$ is below their baseline contribute to the pseudo-label for that sample.

\textit{Step 3}: each selected model $w_m \in \mathbb{M}(x)$ contributes a prediction vector $\mathbf{y}_{m}$.
Inspired by negative learning \cite{Kim2019}, which emphasizes not only endorsing correct labels but also actively rejecting incorrect ones, $\mathbf{y}_{m}$ is designed to explicitly support the predicted class while rejecting alternatives.:
\begin{equation}
y_{m, c} \gets 1 \text{, if } c = \arg \max_{c'} \:w_m(x)_{c'}  \text{ else } -1
\end{equation}

\textit{Step 4}: The final aggregated prediction for $x$ is computed as a weighted vote, when $\mathbb{M}(x) \neq \emptyset$:
\begin{equation}
\label{eq:G}
\mathcal{G}(w_s, x, \rho) = \frac{\sum_{m \in \mathbb{M}(x)} \mathbf{C^2}_{m} \cdot \mathbf{y}_{m}}{\sum_{m \in \mathbb{M}(x)} \mathbf{C^2}_{m}}
\end{equation}
%
where $\mathbf{C^2}_{m}$ denotes the class-wise confidence scores for model $w_m$, reflecting its relative strength in each class.

\textit{Step 5}: 
Based on the aggregated result $\mathcal{G}(w_s, x, \rho)$, we generate the pseudo-label by the one-hot encoding function $\phi(\cdot)$.
By systematically aggregating these weighted votes, the generated pseudo-label is a comprehensive synthesis of all models' inputs, reflecting a balanced classification.

\noindent\textbf{Server model training:}
We optimize the server model by jointly minimizing: the pseudo-label loss $\mathcal{L}_{u}(w_s, \hat{\mathbf{y}})$ and the distillation loss $\mathcal{L}_{d}(w_s)$ from the client model predictions.
As in the objective function (\ref{eq:overall}), our approach evaluates discrepancies for each client model separately and aggregates them with weights.
This strategy effectively captures each client model's unique contributions, in contrast to existing methods that combine all client outputs before learning.

As highlighted before, a critical component of $\mathcal{L}_{d}(w_s)$ is the weight $\lambda_{k}(x)$, which measures the similarity between the unlabeled sample $x$ and the private data $\mathbb{D}_{l, k}$ of the client $k$.
Recognizing that a model's prediction confidence reflects its training data characteristics, we define $\lambda_{k}(x)$ based on prediction confidence, similar to FedHKT \cite{Deng2023}, as follows:
\begin{equation}
\label{eq:weight}
    \lambda_{k}(x) = \frac{\exp(O^{k}(x))}{\sum_{j=1}^{j=K} \exp(O^{j}(x))}
\end{equation}
where $O^{k}(x) = -\mathcal{H}(\sigma(w_c^{k}(x)))$.
This scheme prioritizes client models with lower entropy (i.e., higher confidence), thereby enhancing their influence on the aggregated loss.

\begin{algorithm}[t!]
\caption{Pseudo label generation}
\label{alg:initial}
\begin{algorithmic}[1]
\Input {$\rho$: the participation ratio;
$\mathbb{D}_{u}=\{{x}\}$: unlabeled public dataset; 
$\{w_m(x)\}$: the prediction outputs for each source model $w_m \in \mathbb{M}(x)$;
$\mathbf{C^2}_{m}$: the class-wise confidence scores for each model $w_m$; 
}

\Output{$\{\hat{\mathbf{y}}\}$}, pseudo labels for $\mathbb{D}_{u}$.

\State Generate the entropy baseline for each source model $w_m$: $\tilde{H}_m \gets$ $\lceil \rho \: N_u \rceil$-th lowest entropy from $w_m(\mathbb{D}_{u})$,  
 
\For{$x \in \mathbb{D}_{u}$} 
    \State {Initialize $\hat{\mathbf{y}} \gets \mathbf{0}$, $\mathbb{M}(x) \gets \emptyset$}
    \For{$w_m$} 
        \If{$\mathcal{H}(\sigma(w_{m}(x))) \leq \tilde{H}_m$}
            \State  $\mathbb{M}(x) \gets \mathbb{M}(x) \cup \{w_m\}$
            \State Generate label vector $\mathbf{y}_{m} \text{, where}$ 
            \Statex \phantom{ \:\:\:\: \algorithmicindent} $y_{m, c} \gets 1 \text{, if } c = \arg \max_{c'} \:w_m(x)_{c'} \text{ else } -1$
        \EndIf
        
    \EndFor

    \If{$\mathbb{M}(x) \neq \emptyset$}
        \State $\mathcal{G}(w_s, x, \rho) \gets \frac{\sum_{m \in \mathbb{M}(x)} \mathbf{C^2}_{m} \cdot \mathbf{y}_m}{\sum_{m \in \mathbb{M}(x)} {\mathbf{C^2}_{m}}} $
        \State Generate label vector $\hat{\mathbf{y}} \gets \phi(\mathcal{G}(w_s, x, \rho))$
    \EndIf
\EndFor

\end{algorithmic}
\end{algorithm}

\section{NUMERICAL EVALUATION}
\label{sec:eva}


\subsection{Experiment Setup}
\noindent\textbf{Dataset settings:}
For evaluation, we conduct image classification tasks on CIFAR-100, a dataset of $60,000$ color images across $100$ distinct classes ($600$ images per class).
We randomly split the dataset into a public set and a private set. 
The public dataset ($5,000$ samples) has all labels removed and is shared among all participants.
To simulate label skews in the private set, we adopt two methods \cite{Li2022b}: 
1) Dirichlet partition $p \sim \text{Dir}(\alpha)$, where the proportion $p$ of samples for each class allocated to a client is drawn from a Dirichlet distribution (with lower $\alpha$ indicating greater heterogeneity);
2) Pathological partition $\# I_{c}$, where each client receives data exclusively from $I_c$ classes, with an equal number of samples for each class distributed among the assigned clients.

\noindent\textbf{Baseline:}
We evaluate FedOL against the following baseline methods:
1) Local: clients independently train the model on their private dataset without collaboration; 
2) FedAvg \cite{McMahan2017}: a parameter-based method that averages client model parameters;
3) FedProx \cite{Li2020}: a federated optimization algorithm for heterogeneous data;
4) FedDF \cite{Lin2020}: a robust model fusion method over heterogeneous client models;
5) FedHKT \cite{Deng2023}: a hierarchical knowledge transfer framework;
6) FedET \cite{Cho2022}: an ensemble knowledge transfer method with customized models;
7) MinE \cite{Grandvalet2004}: a voting method that chooses the prediction with the lowest entropy; 
8) FedKT \cite{Li2021}: a one-shot FL method by utilizing knowledge transfer.
For fairness, we exclude methods requiring a labeled public dataset (e.g., FedGEMS \cite{Cheng2021}) or those generating labeled data (e.g., DENSE \cite{Zhang2022}).
For further details of each baseline, please refer to the corresponding papers.

\noindent\textbf{Model settings:} 
For parameter-based methods, all the clients and the server use the ResNet20 model;
For knowledge-based methods, clients use the ResNet11 or ResNet20 model, and the server side utilizes a larger ResNet56 model. 

\noindent\textbf{Parameters:}
By default, the system includes $K = 10$ clients. 
For each model training, we run $50$ epochs, with the batch size being $64$ and the learning rate being $0.001$.
We set the iteration round in FedOL to $10$, loss tradeoff $\tau$ in (\ref{eq:overall}) to $0.2$.
The parameter $\rho$, used in setting the entropy baseline, starts at $0.1$ and increases by $0.05$ with each iteration round.

\subsection{Performance Analysis}

\begin{figure}[t!]
\centering
\begin{subfigure}{.45\linewidth}
\resizebox{\textwidth}{!}{
\begin{tikzpicture}
\begin{axis}[
    xlabel={Communication rounds},
    ylabel={Model accuracy \%},
    ylabel style={at={(axis description cs:0.1,0.75)}, anchor=south}, 
    xmin=0, xmax=15,
    ymin=0, ymax=50,
    xtick={0,2,4,6,8,10,12,14},
    ytick={0,20,40, 60},
    legend style={draw=none},
    legend pos=south east,
    legend columns=2, 
    ymajorgrids=true,
    grid style=dashed,
]

\addplot [
    black, very thick,
    domain=0:15, 
    samples=2, 
] {37.2} node [pos=0.5, above] {}; 
\addlegendentry{FedOL}
        
\addplot[
    color=darkgray,
    mark=square,
    ]
    coordinates {
    (1,1.0)(2, 14.9)(3, 33.1)(4, 38.0)(5, 41.4)(6, 42.0)(7, 43.4)(8, 43.5)(9, 45.6)(10, 45.8)(11, 45.6)(12, 47.0)(13, 46.2)(14, 47.1)(15, 47.4)
    };
    \addlegendentry{FedAvg}

\addplot[
    color=gray,
    mark=|,
    ]
    coordinates {
    (1,1.7)(2,5.6)(3,23.4)(4,30.0)(5,33.3)(6,37.3)(7,37.7)(8,38.7)(9,39.3)(10,37.6)(11,38.9)(12,40.0)(13,38.8)(14,38.8)(15,40.7)
    };
    \addlegendentry{FedProx}

\addplot[
    color=darkgray,
    mark=10-pointed star,
    ]
    coordinates {
    (1,29.8)(2, 31.0)(3, 31.8)(4, 31.6)(5, 30.8)(6, 31.0)(7, 31.6)(8, 32.0)(9, 31.6)(10, 31.2)(11, 30.2)(12, 29.5)(13, 31.4)(14, 31.2)(15, 30.4)
    };
    \addlegendentry{FedDF}

\addplot[
    color=gray,
    mark=triangle,
    ]
    coordinates {
    (1,29.4)(2, 30.4)(3, 32.3)(4, 32.3)(5, 33.1)(6, 33.2)(7, 33.5)(8, 33.3)(9, 33.5)(10, 31.7)(11, 32.4)(12, 32.4)(13, 32.3)(14, 31.0)(15, 33.3)
    };
    \addlegendentry{FedHKT}
        
\addplot[
    color=darkgray,
    mark=x,
    ]
    coordinates {
    (1,25.8)(2,28.7)(3,28.1)(4,28.5)(5,27.8)(6,26.7)(7,26.5)(8,29.3)(9,29.2)(10,30.6)(11,33.1)(12,33.3)(13,32.7)(14,34.8)(15,35.1)
    };
    \addlegendentry{FedET}
        
\addplot[
    color=gray,
    mark=o,
    ]
    coordinates {
    (1,26.2)(2, 27.4)(3, 28.7)(4, 28.7)(5, 27.9)(6, 29.6)(7, 28.7)(8, 27.3)(9, 27.9)(10, 28.9)(11, 28.5)(12, 27.7)(13, 29.7)(14, 29.2)(15, 29.3)
    };
    \addlegendentry{MinE}
    
\end{axis}
\end{tikzpicture}
}
\caption{$p \sim \text{Dir}(1)$}
\label{fig:100dir1}
\end{subfigure}%
\hspace{0.25em}
\begin{subfigure}{.45\linewidth}
\resizebox{\textwidth}{!}{
\begin{tikzpicture}
\begin{axis}[
    xlabel={Communication rounds},
    ylabel={Model accuracy \%},
    ylabel style={at={(axis description cs:0.1,0.75)}, anchor=south}, 
    xmin=0, xmax=15,
    ymin=0, ymax=40,
    xtick={0,2,4,6,8,10,12,14},
    ytick={0,20,40},
    legend style={draw=none},
    legend pos=south east,
    legend columns=2, 
    ymajorgrids=true,
    grid style=dashed,
]

\addplot [
    black, very thick,
    domain=0:15, 
    samples=2, 
] {29.8} node [pos=0.5, above] {}; 

\addplot[
    color=darkgray,
    mark=square,
    ]
    coordinates {
    (1, 1.0)(2, 3.8)(3, 12.0)(4, 17.8)(5, 22.8)(6, 26.4)(7, 27.3)(8, 30.1)(9, 30.9)(10, 31.6)(11, 32.6)(12, 35.4)(13, 35.0)(14, 35.6)(15, 36.3)
    };

\addplot[
    color=gray,
    mark=|,
    ]
    coordinates {
    (1,1.0)(2,2)(3,2.9)(4,5.1)(5,8)(6,10.8)(7,14.4)(8,18.7)(9,18.7)(10,20.9)(11,23)(12,22.8)(13,24.4)(14,24.2)(15,24.6)
    };
        
\addplot[
    color=darkgray,
    mark=10-pointed star,
    ]
    coordinates {
    (1,14.8)(2, 17.6)(3, 19.0)(4, 20.1)(5, 20.9)(6, 20.1)(7, 20.1)(8, 21.9)(9, 21.5)(10, 20.4)(11, 20.5)(12, 21.2)(13, 21.7)(14, 22.0)(15, 21.3)
    };

\addplot[
    color=gray,
    mark=triangle,
    ]
    coordinates {
    (1,18.8)(2, 20.0)(3, 22.5)(4, 23.8)(5, 24.5)(6, 24.4)(7, 24.4)(8, 24.2)(9, 24.8)(10, 24.3)(11, 25.0)(12, 24.5)(13, 24.7)(14, 25.5)(15, 25.1)
    };
    
\addplot[
    color=darkgray,
    mark=x,
    ]
    coordinates {
    (1,18.0)(2,18.8)(3,18.1)(4,20)(5, 20.5)(6, 21.4)(7, 21.8)(8, 22.1)(9, 22.7)(10, 22.4)(11, 23.2)(12, 23.2)(13, 23.8)(14, 22.6)(15, 23.9)
    };

\addplot[
    color=gray,
    mark=o,
    ]
    coordinates {
    (1,15.5)(2,16.5)(3,17)(4,19)(5,18.5)(6,18.5)(7,18.4)(8,18.6)(9,17.8)(10,17.4)(11,18.2)(12,17.7)(13,17.2)(14,18)(15,17.5)
    };
    
\end{axis}
\end{tikzpicture}
}
\caption{$p \sim \text{Dir}(0.05)$}
\label{fig:100dir005}
\end{subfigure}%
\\
\begin{subfigure}{.45\linewidth}
\resizebox{\textwidth}{!}{
\begin{tikzpicture}
\begin{axis}[
    xlabel={Communication rounds},
    ylabel={Model accuracy \%},
    ylabel style={at={(axis description cs:0.1,0.75)}, anchor=south}, 
    xmin=0, xmax=15,
    ymin=0, ymax=50,
    xtick={0,2,4,6,8,10,12,14},
    ytick={0,20,40},
    legend style={draw=none},
    legend pos=south east,
    legend columns=2, 
    ymajorgrids=true,
    grid style=dashed,
]

\addplot [
    black, very thick,
    domain=0:15, 
    samples=2, 
] {35.8} node [pos=0.5, above] {}; 

\addplot[
    color=darkgray,
    mark=square,
    ]
    coordinates {
    (1,1.1)(2,1.3)(3,3.5)(4,17.1)(5,27.0)(6,29.9)(7,35.4)(8,37.5)(9,37.2)(10,39.2)(11,40.3)(12,42.3)(13,43.0)(14,44.0)(15,42.5)
    };
        
\addplot[
    color=gray,
    mark=|,
    ]
    coordinates {
    (1,1.5)(2, 1.7)(3, 7.8)(4, 13.4)(5, 21.1)(6, 22.6)(7, 28.0)(8, 29.5)(9, 31.1)(10, 31.9)(11, 35.3)(12, 33.5)(13, 36.2)(14, 37.0)(15, 38.2)
    };

\addplot[
    color=darkgray,
    mark=10-pointed star,
    ]
    coordinates {
    (1,26.1)(2,23.9)(3,22.2)(4,27.4)(5,29.7)(6,30.5)(7,31.6)(8,32.8)(9,30.3)(10,30.0)(11,29.7)(12,31.2)(13,30.2)(14,31.7)(15,30.7)
    };

\addplot[
    color=gray,
    mark=triangle,
    ]
    coordinates {
    (1,26.2)(2,29.3)(3,30.0)(4,28.0)(5,26.8)(6,28.8)(7,30.8)(8,30.6)(9,31.1)(10,30.5)(11,28.8)(12,31.0)(13,30.7)(14,30.1)(15,30.7)
    };
    
\addplot[
    color=darkgray,
    mark=x,
    ]
    coordinates {
    (1,25.0)(2, 25.1)(3, 25.7)(4, 25.2)(5, 25.6)(6, 25.2)(7, 25.1)(8, 26.7)(9, 25.4)(10, 25.9)(11, 25.9)(12, 26.2)(13, 26.8)(14, 27.2)(15, 28.0)
    };
        
\addplot[
    color=gray,
    mark=o,
    ]
    coordinates {
    (1,30.0)(2,31.7)(3,32.7)(4,32.1)(5,32.4)(6,32.6)(7,31.5)(8,31.4)(9,30.6)(10,32.4)(11,32.2)(12,31.1)(13,31.0)(14,30.3)(15,31.5)
    };

\end{axis}
\end{tikzpicture}
}
\caption{$ \# I_c = 30$}
\label{fig:100class30}
\end{subfigure}%
\hspace{0.25em}
\begin{subfigure}{.45\linewidth}
\resizebox{\textwidth}{!}{
\begin{tikzpicture}
\begin{axis}[
    xlabel={Communication rounds},
    ylabel={Model accuracy \%},
    ylabel style={at={(axis description cs:0.1,0.75)}, anchor=south}, 
    xmin=0, xmax=15,
    ymin=0, ymax=40,
    xtick={0,2,4,6,8,10,12,14},
    ytick={0,20,40},
    legend style={draw=none},
    legend pos=south east,
    legend columns=2, 
    ymajorgrids=true,
    grid style=dashed,
]

\addplot [
    black, very thick,
    domain=0:15, 
    samples=2, 
] {33.5} node [pos=0.5, above] {}; 

\addplot[
    color=darkgray,
    mark=square,
    ]
    coordinates {
    (1,1.0)(2,1.1)(3,2.2)(4,9.2)(5,17)(6,20.8)(7,24.5)(8,27.7)(9,30.8)(10,31.2)(11,32.9)(12,32.4)(13,34.7)(14,34.1)(15,34.9)
    };

\addplot[
    color=gray,
    mark=|,
    ]
    coordinates {
    (1,1.1)(2,1.7)(3,4.7)(4,9.9)(5,11.1)(6,12.6)(7,17.4)(8,19.7)(9,22.3)(10,23.7)(11,23.4)(12,26)(13,27.9)(14,27.2)(15,26)
    };
        
\addplot[
    color=darkgray,
    mark=10-pointed star,
    ]
    coordinates {
    (1,21.3)(2,21.2)(3,21.6)(4,20.0)(5,21.1)(6,22.8)(7,22.8)(8,20.2)(9,21.5)(10,22.9)(11,21.7)(12,22.1)(13,18.9)(14,20.3)(15,21.5)
    };

\addplot[
    color=gray,
    mark=triangle,
    ]
    coordinates {
    (1,25.2)(2,20.5)(3,24.4)(4,24.6)(5,25.4)(6,23.9)(7,22.7)(8,20.6)(9,20.5)(10,22)(11,23.2)(12,24.5)(13,24.7)(14,25.5)(15,24.7)
    };
    
\addplot[
    color=darkgray,
    mark=x,
    ]
    coordinates {
    (1,22.2)(2,23.7)(3,24.9)(4,23)(5,23.1)(6,24)(7,23.1)(8,22.7)(9,22.7)(10,23.1)(11,23.7)(12,24.5)(13,24.7)(14,25.5)(15,25.6)
    };

\addplot[
    color=gray,
    mark=o,
    ]
    coordinates {
    (1,17.5)(2,15.4)(3,19.9)(4,21.3)(5,19.3)(6,21)(7,21.9)(8,20.2)(9,20.2)(10,19.9)(11,21.2)(12,17)(13,20)(14,17.5)(15,17.9)
    };
    
\end{axis}
\end{tikzpicture}
}
\caption{$ \# I_c = 20$}
\label{fig:100class20}
\end{subfigure}%
\caption {
Performance under multiple communication rounds.
}
\label{fig:cifar100}
\end{figure}
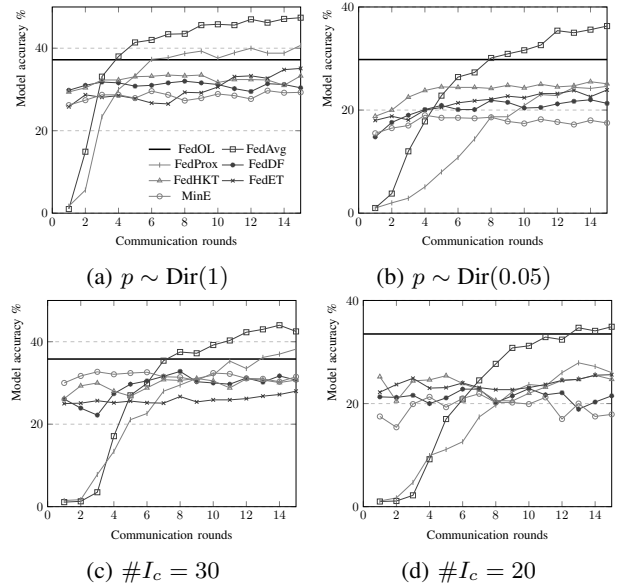

\renewcommand{\arraystretch}{0.2}
\begin{table}[t!]
\centering
\caption{Cost per client in one communication round}
\label{tab:cost}
\begin{adjustbox}{width=0.48\textwidth}
\begin{tabular}{c|c| c c |c c c }
\toprule
\textbf{Type} & \textbf{FedOL} & \textbf{Local} & \textbf{FedAvg} & \textbf{FedDF} & \textbf{FedET} & \textbf{FedKT} \\
\midrule 
Comm/MB & 0.38 & \multicolumn{2}{|c|}{65.14} & \multicolumn{3}{c}{0.38} \\
\midrule
Comp & \multicolumn{6}{c}{standard supervised‐learning procedure} \\
\bottomrule 
\end{tabular}
\end{adjustbox}
\end{table}

\noindent\textbf{Comparison under one communication round:}
For a fair comparison aligned with practical model market usage, all methods are evaluated in a single communication round, as shown in Table~\ref{tab:Diri}. 
Across every partition setting, FedOL outperforms the best baseline by over $5.8\%$, underscoring its robustness in extracting diverse client knowledge. 
This advantage is particularly evident under severe label skew (e.g., $p \sim \text{Dir}(0.05)$ or $\# I_c = 20$), where both parameter-based and knowledge-based baselines degrade significantly. 
By contrast, FedOL effectively leverages heterogeneous client predictions and mitigates skew-related issues, leading to better generalization despite imbalanced private data. 

\noindent\textbf{Comparison under multi-communication round:}
In this simulation, baseline methods (2–7) each run for $15$ communication rounds, in contrast to our FedOL operating in only one communication round. 
The multi-round setting allows the server to provide feedback, letting clients refine and resend information repeatedly after the initial sharing. 
As illustrated in Fig.~\ref{fig:cifar100}, while the baselines gradually improve accuracy over multiple rounds, FedOL rapidly achieves substantial performance with just a single round. 
Notably, it consistently matches or surpasses knowledge-based approaches (4–7) even after their $15$ rounds. 
Similarly, FedOL shows strong initial accuracy compared to parameter-based methods like FedAvg and FedProx, which require extensive tuning to achieve better performance. 
Across all scenarios, FedAvg requires \textit{an average of $10$ communication rounds} to approach FedOL's performance, leading to significantly higher communication and computation costs.
These findings underscore FedOL’s efficiency and potential for quick, low-overhead deployments.

\subsection{Cost Analysis}
We compare the cost per client, with the findings for one communication round summarized in Table~\ref{tab:cost}.
Communication cost is evaluated based on the data transmitted from each client to the server. 
For methods utilizing parameter-based sharing, a ResNet20 model ($17$ million parameters at $4$ bytes each) results in a transmission size of $65.14$ MB.
In contrast, knowledge-based methods transmit predictions for a public dataset of $5000$ samples, each as a probability distribution over $10$ classes, resulting in a total data size of only $0.38$ MB.
All methods employ the same standard supervised‐learning procedure on the client side and therefore incur identical computational costs.
In multi-round scenarios, \textbf{both communication and computation costs increase with the number of rounds}, often exceeding linear scaling. 
Overall, our method achieves the lowest computation and communication costs while maintaining superior performance.

\section{CONCLUSION}
In mobile networks, servers typically possess abundant computational resources, whereas decentralized clients have vast amounts of data but limited computational and communication capabilities. 
To appropriately allocate tasks according to these varying capacities without leakage of privacy, we propose FedOL, a knowledge-based FL algorithm that builds a larger and generalized server model by asking clients share their model predictions just once.
This one-shot approach reduces computational demands on clients and enhances data privacy by minimizing data exposure compared to traditional multi-round FL methods.
Moreover, we introduce novel pseudo-label generation and model refinement techniques to address biased client predictions and the lack of ground truth in the unlabeled dataset.
Extensive simulations demonstrate that FedOL outperforms existing FL algorithms in both performance and cost efficiency, offering promising directions for advancing distributed AI in mobile networks.

\end{document}